\documentclass[final]{cvpr}

\usepackage{times}
\usepackage{epsfig}
\usepackage{graphicx}
\usepackage{amsmath}
\usepackage{amssymb}
\usepackage{xcolor}
\definecolor{asparagus}{rgb}{0.53, 0.66, 0.42}
\usepackage{bm}
\usepackage{enumitem}
\usepackage{nopageno}

\usepackage[pagebackref=true,breaklinks=true,colorlinks,bookmarks=false]{hyperref}

\newcommand{\ucns}{cross-lingual cross-modal pre-training}
\newcommand{\ucabbr}{\textsc{UC$^2$} }
\newcommand{\ucabbrns}{\textsc{UC$^2$}}

\newcommand{\head}[1]{\noindent\textbf{#1}}

\def\cvprPaperID{1719} 
\def\confYear{CVPR 2021}

\begin{document}

\title{\ucabbrns: Universal Cross-lingual Cross-modal Vision-and-Language Pre-training}
\author{Mingyang Zhou$^1$, Luowei Zhou$^2$, Shuohang Wang$^2$, Yu Cheng$^2$, Linjie Li$^2$, Zhou Yu$^1$, Jingjing Liu$^2$ \\
$^1$University of California, Davis \\
$^2$Microsoft Dynamics 365 AI Research \\
{\tt\small \{minzhou, joyu\}@ucdavis.edu} \\
{\tt\small \{luozhou, shuowa, yu.cheng, lindsey.li, jingjl\}@microsoft.com}
}

\maketitle

\begin{abstract}
  Vision-and-language pre-training has achieved impressive success in learning multimodal representations between vision and language. To generalize this success to non-English languages, we introduce UC$^2$, the first machine translation-augmented framework for cross-lingual cross-modal representation learning. 
  To tackle the scarcity problem of multilingual captions for image datasets, we first augment existing English-only datasets with other languages via machine translation (MT).
  Then we extend the standard Masked Language Modeling and Image-Text Matching training objectives to multilingual setting, where alignment between different languages is captured through shared visual context (\ie, using image as pivot). To facilitate the learning of a joint embedding space of images and all languages of interest, we further propose two novel pre-training tasks, namely Masked Region-to-Token Modeling (MRTM) and Visual Translation Language Modeling (VTLM), leveraging MT-enhanced translated data.
  Evaluation on multilingual image-text retrieval and multilingual visual question answering benchmarks demonstrates that our proposed framework achieves new state of the art on diverse non-English benchmarks while maintaining comparable performance to monolingual pre-trained models on English tasks.
   
\end{abstract}

\section{Introduction}
The world we navigate through is a multimodal and multilingual kaleidoscope. While tremendous success has been realized in multimodal research with the advent of vision-and-language (V+L) pre-training~\cite{UNITER,OSCAR,vilbert,tan2019lxmert, huang2020pixelbert}, the majority of current literature is biased towards English. Although English-trained V+L models can be finetuned on each target language (given that there is sufficient language-specific data in downstream task), maintaining language-specific models for every language in the world (6,900+) is impossible given insurmountable development and maintenance cost \cite{hu2020xtreme}. 
Naturally, a ``Tower of Babel'' strategy starts to gain interest in the community, aiming at building one giant model that can handle all languages, notable examples including massively multilingual neural machine translation~\cite{aharoni2019massively}, cross-lingual language model~\cite{lample2019cross}, and multilingual multimodal representation learning~\cite{gella2017image,huang2020m3p}.

\begin{figure}[t!]
\centering
\includegraphics[width=0.9\linewidth]{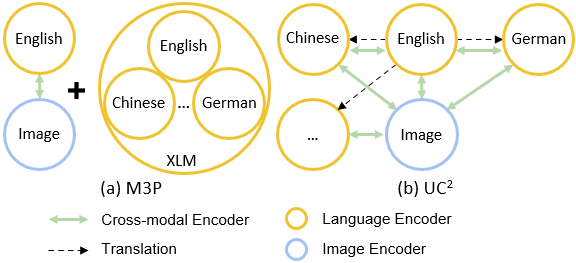}
  \caption{ A topology comparison between existing work (M3P) and our proposed \ucabbrns. 
  M3P combines two types of pre-training tasks and the cross-modal Transformer works only on images and English captions. Our UC$^2$ builds a cross-lingual cross-modal Transformer over images and all the other languages.
  }
\label{fig:comp}
\end{figure}

\begin{figure*}[h!]
\centering
\includegraphics[width=16cm]{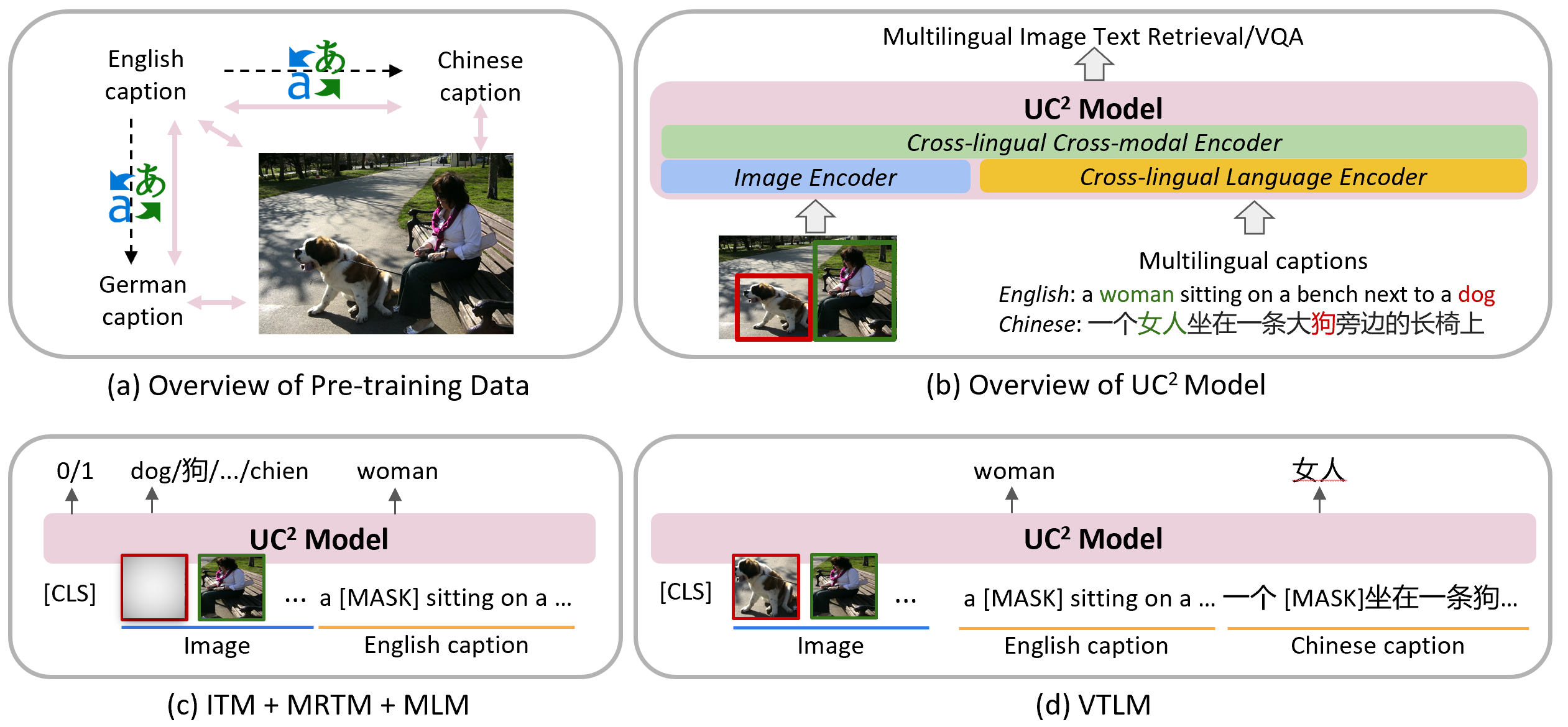}
\caption{An overview of UC$^2$ model. Figure (a) shows the construction of multilingual multimodal pre-training corpus via machine translation. 
(b) depicts the overall UC$^2$ framework, which is pre-trained with a massive corpus of multilingual caption-image pairs. 
Figure (c) and (d) illustrate details of four pre-training tasks.}

\label{fig:model}
\end{figure*}

Early works on cross-lingual multimodal tasks mainly focus on machine translation~\cite{hewitt2018learning,zhou2018visual,calixto-liu-2017-incorporating,yao-wan-2020-multimodal,DBLP:journals/corr/abs-1807-11605} and image-text retrieval~\cite{gella2017image,MULE,SMALR,Par.EMnb, S-LIWE}. The goal is to construct a common embedding space for vision and cross-lingual inputs, and draw visual concepts from images and similar semantics from languages close together in the feature space. 
However, due to the scarcity of large-scale training corpora, these models are validated only on small task-specific datasets, thus scaling and generalizing these models to more languages is non-trivial. 

Recent release of large-scale multimodal datasets ~\cite{conceptual-caption} and multilingual corpora (\eg Wikipedia in 100 languages) has served as a key impetus to accelerate fast advances in V+L pre-training ~\cite{UNITER,vilbert,tan2019lxmert,zhou2019vlp} and multilingual language modeling~\cite{conneau2019cross,XLMR,hu2020xtreme}, which makes pre-training large-scale multilingual V+L models possible. A pioneering work is $\text{M}^3\text{P}$~\cite{huang2020m3p}, which formulates the training process as alternating V+L pre-training between cross-modal monolingual corpus and mono-modal cross-lingual corpus. 
It relies on English as the focal point to build a bridge between image and different languages, which inevitably introduces linguistic discrepancy into downstream tasks that rely on direct alignment between image and Non-English language (\eg, Image-to-German retrieval), as shown in Figure~\ref{fig:comp} (a).

In this paper, we propose a new pre-training framework, UC$^2$ (\textbf{U}niversal \textbf{C}ross-lingual \textbf{C}ross-modal pre-training), which pivots primarily on images and complementarily on English for multilingual multimodal representation learning (Figure~\ref{fig:comp} (b)). The major challenge is that pivoting on images requires paired image and aligned multilingual data (\eg, image-English, image-German), while existing V+L datasets only contain image-English pairs.
To fill this blank, we propose to augment English-only datasets with other languages via machine translation (MT), and leverage the augmented datasets for pre-training.
To the best of our knowledge, this is the first known effort in creating large-scale training datasets with multilingual image captions.

In addition to extending two widely-adopted pre-training tasks (Masked Language Modeling and Image-Text Matching) to a multilingual setting, we further propose two novel pre-training objectives, namely Masked Region-to-Token Language Modeling (MRTM) and Visual Translation Language  Modeling (VTLM). MRTM encourages fine-grained alignment between words and image regions, by sharing the embedding space of word tokens and region labels (\ie, object class predictions from an object detector). VTLM is designed to jointly learn cross-lingual cross-modal mapping from parallel textual corpora and paired images. Extensive experiments demonstrate that our proposed UC$^2$ framework achieves new state of the art over multiple mainstream benchmarks such as Multi30k~\cite{multi30k, multi30k_extension_1, multi30k_extension_2} and COCO~\cite{MSCOCO, MSCOCO_JA, MSCOCO_ZH} across multilingual image-text retrieval and visual question answering (VQA) tasks.

Our contributions are summarized as follows. $(i)$ We construct a multilingual V+L corpus, and propose the first MT-augmented cross-lingual cross-modal pre-training framework UC$^2$, which pivots on both images and English language for joint representation learning.
$(ii)$ We propose new pre-training tasks, Masked Region-to-Token Language Modeling and Visual Translation Language Modeling, two effective learning objectives for multilingual multimodal tasks. 
$(iv)$ We achieve new state of the art on multiple multilingual image-text retrieval and VQA benchmarks, outperforming existing methods. 
\section{Related Work}
\paragraph{Vision-Language Pre-training.}
There is a growing interest in building generic pre-trained BERT-like~\cite{BERT} models for V+L tasks. Early work such as VilBERT \cite{vilbert} and LXMERT \cite{tan2019lxmert} propose a two-stream architecture that encodes visual and textual input through two separate Transformers, and then fuse the two modalities by a cross-modal Transformer. Later work such as VL-BERT \cite{su2020vlbert}, Unicoder-VL\cite{unicodervl} and UNITER \cite{UNITER} introduce a single-stream architecture that uses one Transformer to encode concatenated input from both modalities simultaneously. Later, Unified VLP \cite{zhou2019vlp} applies to both understanding and generation tasks. Further improvements are proposed on using different input features \cite{OSCAR, huang2020pixelbert} and multi-task learning \cite{lu202012in1}. 

\vspace{-3mm}
\paragraph{Multimodal Multilingual Learning.} 
Existing studies arching over multilingual and multimodal aspects mainly focus on two tasks: cross-modal retrieval and multimodal machine translation (MT). 
\cite{rajendran-etal-2016-bridge, calixto-liu-2017-sentence} introduces a multimodal multilingual approach by aligning images and captions in  different languages to English captions. Unlike previous work using languages as a pivoting point, \cite{gella2017image} learns a shared embedding space that forces representations of different languages towards the pivot image representation. Later work focuses on scaling to more languages via character-based word-embedding \cite{S-LIWE} or shared language-acoustic embedding \cite{MULE}. SMALR \cite{SMALR} proposes a scalable multilingual model to learn visually aligned word embeddings, for better balance between multilingual capacity and task performance.

Multimodal MT exploits visual information to improve language translations. Earlier work introduces vision to an LSTM-based neural MT model via attention to visual context \cite{calixto-etal-2017-doubly, helcl-libovicky-2017-cuni}, or fusion \cite{calixto-liu-2017-incorporating}, or multi-task learning\ \cite{elliott-kadar-2017-imagination, zhou2018visual}. Lately, Transformer-based~\cite{NIPS2017_3f5ee243} models are proposed~\cite{DBLP:journals/corr/abs-1807-11605, yao-wan-2020-multimodal}. There is also an growing interest in unsupervised multimodal MT \cite{huang-etal-2020-unsupervised,Sigurdsson_2020_CVPR}, where translation between monolingual corpus is augmented via pivoting on image. 


While successful in individual tasks, these models are usually trained on small amount of data, which limits its extension to other tasks or languages. To learn task-agnostic universal representations across vision and multilingual text,  $\text{M}^3\text{P}$\cite{hu2020xtreme} introduces the first pre-training framework that alternatively optimizes the model on multimodal monolingual corpus and mono-modal multilingual corpus. While $\text{M}^3\text{P}$ achieves better performance compared to task-specific methods, the alignment between vision and Non-English languages is hard to capture, as the model is learned via using English as the anchor point. To strengthen the alignment between vision and all languages, we propose to pre-train a unified architecture where sentences in different languages are grounded on shared visual context. 


\section{Cross-Lingual Cross-Modal Pre-training}

In this section, we start with introducing our machine translation augmented dataset that enables large-scale cross-lingual pre-training. We then go over the proposed UC$^2$ model and our designed pre-training objectives for universal representation learning across vision and languages. 



\subsection{Machine Translation Augmented Dataset}
Our multilingual image-text paired data is collected via augmenting the captions
from the Conceptual Captions dataset~\cite{conceptual-caption} with a set of machine translated\footnote{We use Microsoft Azure Translation API Service and will release the translated captions.} captions in other languages $\boldsymbol{L} = \{l_1, l_2, \dots, l_n\}$.
Specifically, we translate the original English captions into five different languages (German, French, Czech, Japanese, and Chinese), which covers languages required for all the downstream tasks studied in this work. Note that with recent advances on machine translation for low-resource languages, we can further expand the dataset to more languages, which we leave for future work. With this data augmentation, we obtained 3.3 million images, each paired with captions in six languages, as the process shown in Figure~\ref{fig:model} (a). 
This one-to-many mapping greatly facilitates the learning of alignment between visual content and semantics from each language through image as a shared anchor. 
By introducing translated data into model pre-training, our method yields significant improvement over the baseline with MT tools applied only on downstream tasks.
Next, we elaborate how to leverage these data for \ucns. 

\subsection{Model Overview}
UC$^2$ extends monolingual language encoder of V+L frameworks, such as UNITER~\cite{UNITER}, to cross-lingual encoder~\cite{XLMR}, as shown in Figure~\ref{fig:model} (b).   
 The visual feature is extracted from an image encoder and the language feature is obtained from a general cross-lingual language encoder.
The multimodal features are then combined into a sequence and fed to a multi-layer Transformer to produce contextualized cross-modal and cross-lingual representations.

\vspace{5pt}
\head{Image Encoder.}
Given an input image, we first obtain a sequence of image region features $\boldsymbol{v} = \{v_1, v_2, \cdots, v_m\}$ with Faster R-CNN \cite{faster-rcnn}. For each region, we also extract location features via a 7-dimensional vector: $\boldsymbol{p} = [x_1, y_1, x_2, y_2, w, h, w*h ]$, which denotes the normalized top left coordinates, bottom right coordinates, width, height, and the area of the detected region box. The region feature and location feature are fed through separate fully-connected (FC) layers to be projected into the same dimension as the text embedding space, followed by a layer-normalization (LN) layer. The final representation of the region feature is then obtained via summing up the projected region feature and location feature.

\vspace{5pt}
\head{Cross-lingual Language Encoder.}
We follow XLM-R \cite{XLMR} to tokenize an input sentence $T^{l_i}$ in language $l_i$ to BPE tokens $\boldsymbol{t^{l_i}} = \{t_1^{l_i}, t_2^{l_i}, \cdots, t_n^{l_i}\}$ using Sentence Piece model \cite{SentencePiece}. We then project each token to its embedding based on the XLM-R vocabulary and word embeddings. 
The final representation of each token is obtained via summing up its word embedding, segment embedding, and position embedding as in XLM-R, followed by another Layer Normalization. 


\subsection{Pre-training Tasks}
For model training, we employ four pre-training objectives to train on large multilingual image-text paired data: Masked Language Modeling (MLM), Image-Text Matching (ITM), Masked Region-to-Token Modeling (MRTM), and Visual Translation Language Modeling (VTLM), as shown in Figure~\ref{fig:model} (c) and (d). We continuously optimize our model with the four objectives on multilingual image-text pairs to capture the cross-modal alignment between vision and different languages. As the translated captions are associated to the same image, cross-lingual alignment is also enforced using visual context as the anchor. 
\subsubsection{General Tasks}
Following previous V+L pre-training work \cite{UNITER, unicodervl, vilbert, su2020vlbert}, we consider Masked Language Modeling and Image-Text Matching as two of our pre-training tasks.

\vspace{5pt}
\head{Masked Language Modeling (MLM).}
 Given a set of image regions $\boldsymbol{v} = \{v_1, v_2, \cdots, v_m\}$ and its associated caption words $\boldsymbol{w^{l_i}} = \{w^{l_ii}_1, \cdots, w^{l_i}_{T}\}$ in language $l_i \in \boldsymbol{L}$, and mask indices as $m \in \mathbb{N}^M$, we randomly mask a word $w^{l_i}_m$ with the probability of $15\%$ and replace the masked word with a special token $\text{[mask]}$. The objective is to predict the masked word $w^{l_i}_m$ based on the surrounding words $w_{\textbackslash m}$ and all image regions $\boldsymbol{v},$ by minimizing the negative log-likelihood: 
\begin{equation*}
    \mathcal{L}_{MLM}(\theta) = - \mathbb{E}_{(w^{l_i}, v)\sim D} \log{P_{\theta}(w^{l_i}_m|w^{l_i}_{\backslash m},v)},
\end{equation*}
where $\theta$ is the learnable parameters. Each pair $(\boldsymbol{w^{l_i}}, \boldsymbol{v})$ is sampled from the whole training set $D$. The caption for each language is sampled with even probability $p = 1/|\boldsymbol{L}|$.

\vspace{5pt}
\head{Image-Text Matching (ITM).}
ITM has been widely used in vision-and-language pre-training~\cite{UNITER, unicodervl,vilbert,su2020vlbert} to learn instance-level alignment between image and sentence.
The output of the special token $\text{[cls]}$ is fed through a FC layer and a sigmoid function to predict a score $s_\theta(w^{l_i}, v)$ between 0 and 1, which predicts whether the input image $\boldsymbol{v}$ and the text input $\boldsymbol{w^{l_i}}$ are semantically matched. During training, we sample positive and negative pairs from the dataset $D$ with equal probability at each step. The negative image-text pair is created by replacing the image or text in a matched pair with a randomly-selected distractor from the same mini-batch. The objective is optimized with binary cross-entropy loss: 
\begin{equation*}
    \begin{split}
        \mathcal{L}_{ITM}(\theta) = & - \mathbb{E}_{(w^{l_i}, v)\sim D} [y \log{s_{\theta}(w^{l_i},v)} \\
        & + (1-y)log(1-s_{\theta}(w^{l_i},v))]
    \end{split}
\end{equation*}
where $y \in \{0, 1\}$ indicates whether the input image-text pair is a positive or negative sample. The deployment of MLM and ITM serves as our base model. Next, we introduce two novel objectives to further enhance cross-lingual cross-modal representation learning.

\subsubsection{Masked Region-to-Token Modeling} 
Now that we have a learning objective for language (MLM), how about the vision counterpart? In existing VLP models, Masked Region Modeling (MRM) serves this purpose by predicting the top-1 or soft object label associated with the masked image region. The de facto approach to harvesting object labels is using the predictions from an off-the-shelf object detector (\eg, Faster R-CNN~\cite{faster-rcnn}). However, there are two limitations with this approach. First, the association between object labels from image and word tokens from text is not well utilized. While salient objects detected in the image are usually mentioned in the paired description, MRM misses this connection as it directly predicts masked image region to an index between 0 and 1600.  
Second, the visual embedding extracted from an object detector can differ significantly from pre-trained word embedding due to different embedding space. Existing methods merely rely on \emph{weak} supervision from pre-training objectives to close the gap between these two disparate embedding spaces.
We argue that a well-aligned embedding space is indispensable for our problem, given its complex multilingual multimodal nature. Therefore, we propose to explicitly learn the correspondence between region and word tokens and tackle the aforementioned issues with two strategies.

\vspace{5pt}

\head{Masked Region-to-Token Modeling (MRTM).} This new objective aims to classify each masked region to its ``pseudo'' object label (e.g., ``dog'', ``cat'', provided by a pre-trained object detector), which is the (sub-word) token\footnote{When multiple tokens are associated to one label word, we randomly select one to decode during pre-training} in our word vocabulary that associates with the original object label.
Compared to the MRM objective from previous work~\cite{vilbert, unicodervl, UNITER}, MRTM leverages additional semantic association between object labels and captions to capture semantic alignment between vision and language. More formally, given an image region $\boldsymbol{v_i} \in \boldsymbol{v}$, we set its probability for being masked out as 15\% (as in ~\cite{BERT}). For each masked region, the region feature vector is either replaced by a zero-initialized vector $v_{m}$ ($90\%$ probability) or remains the same ($10\%$). Then we predict the associated ``pseudo'' object label $c_{v_m}^{l_i}$ on the masked region based on the observation of surrounding image regions $v_{\backslash m}$ and the paired caption $w^{l_i}$ in language $l_i$, by minimizing the negative log-likelihood:
\begin{equation*}
    \mathcal{L}_{MRTM}(\theta) = - \mathbb{E}_{(w^{l_i}, v)\sim D} \log{P_{\theta}(c_{v_m}^{l_i}|w^{l_i},v_{\backslash m})}
\end{equation*}


\head{Early Adaptation (EA).} To address the second limitation and facilitate learning of a joint embedding space between vision and language, we warm up the \textit{image encoder} to make sure the output visual embedding shares the same embedding space as word embeddings.
Specifically, each image region is projected to an image region feature $v_{i} \in \mathbb{R}^{p}$ through the \textit{image encoder}, with the same dimension as the word embedding vector. We then extract the word embedding vectors from XLM-R that correspond to the $k$ object categories $\boldsymbol{c} = \{c_1, c_2, \dots, c_k\}$ defined by the object detector. We compute the cosine similarity between the projected image feature $v_{i}$ with the $k$ word embedding vectors followed by a softmax function, resulting in a normalized distribution $h_{\theta_{I}}(v_i) \in \mathbb{R}^{k}$ that indicates the prediction on what semantics are mapped in the region. We then maximize the similarity between this predicted distribution and the ``GT'' object probability distribution from the object detector output $g(v_i) \in \mathbb{R}^{K}$, by minimizing their  KL divergence:
\begin{equation*}
    \mathcal{L}_{EA}(\theta_{I}) = D_{KL}(g(v_i)||h_{\theta_{I}}(v_i)),
\end{equation*}
where $\theta_I$ is the learnable parameters of the \textit{image Encoder}. 

Note that a recent work named OSCAR~\cite{OSCAR} has made a similar effort to close the visual-textual embedding gap by inserting object tags into the input sequence. Compared to OSCAR \cite{OSCAR}, our method has two advantages. First, it does not rely on object tags for downstream tasks, which might not be applicable for image domains that cannot be well covered by the object categories from the pre-trained detector. Second, by forcing image representation to be similar to language representation with EA, our pre-training model can better leverage the initialized weights from language-only pre-trained model to adapt to image modality.


\subsubsection{Visual Translation Language Modeling}
All the objectives mentioned so far operate on image and \emph{monolingual} input, without considering cross-lingual objectives. The correspondence between languages is vital for cross-lingual generalization, clearly observed from existing work on language understanding~\cite{XLMR}. Our proposed methods so far unexceptionally learn cross-lingual correspondence \emph{indirectly} through the image focal point, which might not be sufficient.  We hence propose visual translation language modeling (VTLM), which directly and jointly learns the alignment between visual context and text in different languages.

In VTLM, given an image $\boldsymbol{v}$ and a pair of captions $(w^{l_i}, w^{l_j})$ in two different languages, the goal is to predict masked caption tokens from both languages. One of the two languages is always English, as English captions in our pre-training data are directly from~\cite{conceptual-caption}, while captions in other languages are translated by MT, therefore less reliable. Under this bilingual framework, model input size only grows linearly with more languages.

Besides, as our model is initialized with the weights of a powerful pre-trained multilingual model, it has already learned a good alignment between different linguistic words to some extent. Applying random masking strategy in VTLM is sub-optimal, as  
the model can make a correct prediction by simply translating words from one language to another, without taking into account the visual information from image. 
To encourage the model to fully consider visual context, we introduce a strategy called \emph{co-masking}, where we simultaneously mask out tokens with similar semantic meanings from paired captions to prevent easy translations.

There are a few steps in \emph{co-masking}. First, we apply Fast Align \cite{fast_align} to learn the word alignment between two different languages $(l_i, l_j)$ from the noisy parallel corpus that was created using machine translation. Then, during the pre-training stage, we follow the same strategy as in MLM to randomly mask a token $w^{l_i}_{m}$ from the caption of one language. For the paired caption in the other language $l_j$, we mask the aligned word tokens $w^{l_j}_{k}$ that are predicted from Fast Align. \cite{fast_align} 
The final objective is again to predict masked tokens from both languages by minimizing the negative log-likelihood:
\begin{equation*}
    \mathcal{L}_{VTLM}(\theta) = - \mathbb{E}_{(w^{l_i}, w^{l_j}, v)\sim D} \log{P_{\theta}(w^{l_i}_m, w^{l_j}_k|w^{l_i}_{\backslash m}, w^{l_j}_{\backslash k}, v)}
\end{equation*}

\begin{table*}[!]
\centering
\small
\begin{tabular}{lcccccccc}
\hline
& \multicolumn{4}{c}{Flickr30K} & \multicolumn{3}{c}{MSCOCO} & \\
Method  & EN  & DE   & FR                       & CS                         & EN                            & ZH                          & JA                            & Meta-Ave                          \\ \hline
\multicolumn{9}{l}{\textit{SOTA without pre-training}}\\
\hline
EmbN\cite{EMnb} & 72.0 & 60.3 & 54.8 & 46.3 & 76.8 & 73.2 & 73.5 & 65.3\\
PAR.EmbN \cite{Par.EMnb} & 69.0 & 62.6 & 60.6 & 54.1 & 78.3 & 76.0 & 74.8 & 67.9\\
S-LIWE \cite{S-LIWE} & 76.3 & 72.1 & 63.4 & 59.4 & 80.9 & 73.6 & 70.0 & 70.8\\
MULE \cite{MULE} & 70.3 & 64.1 & 62.3 & 57.7 & {\color[HTML]{0000FF}79.0} & {\color[HTML]{0000FF}75.9} & {\color[HTML]{0000FF}75.6} & 69.3 \\
SMALR \cite{SMALR} & 74.5 & 69.8 & 65.9 & 64.8 & {\color[HTML]{0000FF}81.5} & {\color[HTML]{0000FF}77.5} & {\color[HTML]{0000FF}76.7} & 73.0 \\
\hline
\multicolumn{9}{l}{\textit{English-only Fine-tune}}\\
\hline
$\text{M}^3\text{P}$\cite{huang2020m3p}                & 87.4                                   & 58.5                                 & 46.0                               & 36.8                                 & 88.6                                 & 53.8                               & 56.0                                 & 60.7                                \\ \hline
UC$^2$   & 87.2 & 74.9 & 74 & 67.9                                 & 88.1 & 82 & 71.7 & 78.0 \\ \hline
\multicolumn{9}{l}{\textit{Translate-Test}} \\
\hline
$\text{UNITER}_{\text{CC}}$\cite{UNITER} & 87.7  & 81.2  & 81.9  & 80.2 & 88.4 & 87.3 & 82.2 & 84.1  \\ \hline
\multicolumn{9}{l}{\textit{Single-Language Fine-tune}}\\
\hline
$\text{M}^3\text{P}$\cite{huang2020m3p}               & 87.4                                  & 82.1                                 & 67.3                               & 65.0                                 & 88.6                                 &75.8                              &  80.1                              & 78.0                                 \\ \hline
UC$^2$   &  87.2 &  83.8 &  77.6 &  74.2                                 & 88.1 &  84.9 & 87.3 & 83.3 \\
\hline
\multicolumn{9}{l}{\textit{All-Language Fine-tune}} \\
\hline
$\text{M}^3\text{P}$\cite{huang2020m3p}               & 87.7                                  & 82.7                                 & 73.9                               & 72.2                                 & \textbf{88.7}                                 & 86.2                              &  \textbf{87.9}                              & 82.8                                 \\ \hline
UC$^2$   &  \textbf{88.2} &  \textbf{84.5} &  \textbf{83.9} &  \textbf{81.2}                                 & 88.1 &  \textbf{89.8} & 87.5 & \textbf{86.2} \\ 
\hline
\end{tabular}
\caption{Evaluation results on image-text retrieval over Flickr30K and MSCOCO datasets across different languages. We highlight the MSCOCO results for MULE and SMALR in {\color[HTML]{0000FF}blue} as they are using different dev/test splits of MSCOCO compared to other models.} 
\label{tab:itm}
\vspace{-3mm}
\end{table*}

\section{Experiments}
In this section, 
we provide detailed experiments to evaluate our proposed UC$^2$ model over multilingual image-text retrieval and multilingual VQA tasks. 


\paragraph{Multilingual Image-Text Retrieval}
In the retrieval task, the model retrieves an image from a set of candidates given a caption in a certain language, or vice versa. We consider two datasets: Multi30K \cite{multi30k, multi30k_extension_1, multi30k_extension_2} and MSCOCO \cite{MSCOCO, MSCOCO_JA, MSCOCO_ZH}. Multi30K is built upon Flickr30K \cite{Flickr30K}, where English captions are manually translated to German, French, and Czech. It contains 31K images (each paired with 5 captions in English and German, 1 caption in French and Czech). Following Flickr30K\cite{Flickr30K}, we split the data into 29K/1K/1K images for train/val/test. 

MSCOCO\cite{MSCOCO} consists of 123K images, with 5 English captions per image. STAIR \cite{MSCOCO_JA} extends MSCOCO dataset by collecting 820K Japanese captions for 165K COCO images. Similarly, Li et al. \cite{MSCOCO_ZH} collect Chinese captions for 20K COCO images with roughly 1 caption per image. We use the train/dev/test splits for English and Japanese defined in \cite{KarpathyF14}, and present results on the 1K test set. For MSCOCO Chinese, we follow the original split as in \cite{MSCOCO_ZH}. We compute Recall@K (recall of top K candidates)  for both image-to-text retrieval and text-to-image retrieval with $K=1, 5, 10$. The average of all these 6 evaluation scores, Average Recall (AR)\cite{huang2020m3p}, is used as the final evaluation metric. 

\paragraph{Multilingual Visual Question Answering (VQA)}
In Multilingual VQA, given an image and a question in a certain language, the model predicts an answer based on the visual context in the image. We evaluate our model on two datasets: VQA v2.0~\cite{vqa_v2} and Japanese Visual Genome (VG) VQA~\cite{vqa_ja}. VQA v2.0 is a widely used benchmark for English VQA task. We follow the official partition to split the dataset and report results on Test-Dev set through the official evaluation server. Following \cite{UNITER}, our training is augmented by running on both the training and validation split of VQA v2.0 as well as the VQA from Visual Genome\cite{vg}. 
Visual Genome VQA Japanese \cite{vqa_ja} expands the VG English VQA dataset \cite{vg} by collecting 793K Japanese question answering pairs on 99K images from VG. We use the train/test split in the original VG VQA to split the data into 61K/30K training/test images. 

We formulate VQA as a multi-label classification problem, where the model predicts answer from the candidate pool.\footnote{We only consider top-3129 frequent answers for VQA v2.0 and top-3000 frequent answers for VQA VG Japanese.}
VQA score~\cite{vqa_v2} is used to compare model predictions against 10 human-annotated answers in VQA v2.0.
On Visual Genome VQA Japanese, which only has one ground-truth answer to each question, we use accuracy and BLEU score as the evaluation metrics.\footnote{BLEU score is used to compute a soft mapping score between the predicted answer and the ground-truth answer, assuming answers with many overlapping words should share similar semantic meaning.}

\paragraph{Implementation Details}
UC$^2$ consists of 12 layers of transformer blocks, where each block has 768 hidden units and 12 self-attention heads. Except for the image encoder, the model is initialized with XLM-R \cite{XLMR}. We run continuous pre-training with MLM, ITM, MRTM and VTLM objectives. We use Adam optimizer \cite{ADAM} with a linear warm-up for the first $5\%$ of training, and set the learning rate to $4e-4$. We use Horovod and NCCL for multi-node communications and apply gradient accumulation (every 3 steps) to reduce multi-GPU communication overheads. The batch-size for pre-training is set as 1024 and the dropout rate is 0.1. Pre-training experiments are conducted on 8 Nvidia V100 GPUs for 30 epochs, which takes 4 days to converge.

\subsection{Experimental Results}
We first compare UC$^2$ to various SOTA with or without pre-training on the two downstream tasks. Then, We conduct ablation experiments to study the effectiveness of MRTM and VTLM, as well as the impact of image pivoting. Finally, we visualize the alignments between visual context and cross-lingual text context learned by our pre-trained UC$^2$ model. 

\subsubsection{Evaluation on Multilingual Retrieval}
We compare UC$^2$ with state-of-the-art methods on image retrieval and text retrieval in two different settings: 
\begin{itemize}[noitemsep]
    \item \textbf{English-only Fine-tune:} We finetune the pre-trained model on just the English training data. 
    \item \textbf{Single-Language Fine-tune:} We finetune the pre-trained model on training data for each target language. 
    \item \textbf{All-Language Fine-tune:} We finetune the pre-trained model on merged training data of all languages. 
\end{itemize}
Besides reporting AR on each language, we also compute the Meta-Ave (average of AR across all languages over two datasets) to reflect the overall performance in this task. Given that we have access to pre-trained machine translation models, we also introduce a strong translate-test baseline $\text{UNITER}_{\text{CC}}$ based on~\cite{UNITER}, which is pre-trained on Conceptual Conception English data and finetuned on English training data in the downstream tasks. By translating the test data from other languages to English, $\text{UNITER}_{\text{CC}}$ can be directly applied for text/image retrieval. Results are summarized in Table~\ref{tab:itm}. 


%

Our model on the all-language setting achieves a significant improvement over all task-specific methods without pre-training, showing the effectiveness of cross-lingual cross-modal  pre-training in learning universal representation across vision and different languages. Our model also demonstrates a superior transferability. When finetuned on English dataset only, we observe an absolute gain of $17.3\%$ on  Meta-Ave
across different languages over M3P via better transition of the learned knowledge from English to other languages.
Compared to the best non-pretrained models trained on data in each language, our cross-lingual model under the English-only fine-tune setting is still $5\%$ better. We suspect the improvement comes from the in-domain pre-training objective: we use image as the grounding media in ITM to learn cross-modal mapping from one language to another. With strong transfer capability, our model could potentially generalize the learned knowledge from a high-resource language to downstream tasks in low-resource languages. 

When we finetune UC${^2}$ model on all-language data, our model still demonstrates a consistent advantage over $\text{M}^3\text{P}$ on the majority of languages, with $3.4\%$ improvement on Meta-Ave. Our best model is also better than the strong translate-test baseline $\text{UNITER}_{\text{CC}}$ on all languages except English in MSCOCO. The slightly worse performance on COCO English is potentially due to lack of pre-training in English data, given that our pre-training time is evenly splitted to multiple languages. However, this does not overshadow the fact that we achieve overall better performance across all languages.  Thanks to the cross-lingual pre-training and finetuning, our model can leverage the complementary information captured in different languages to improve the performance on each language. 

\begin{table}[!]
\small
\centering
\begin{tabular}{l|c|cc}
\hline
& VQA v2.0                              & \multicolumn{2}{c}{VG VQA JA} \\
method & Test-Dev Acc & Acc & BLEU \\
\hline
MCAN \cite{MCAN}                        & 70.63                                & -                                    & -                                    \\ 
PCATT  \cite{vqa_ja}                      & -                                     & 19.2                                 & -                                    \\ \hline
Vil-BERT \cite{vilbert}                    & 70.55                                 & -                                    & -                                    \\ 
VL-BERT \cite{su2020vlbert}                      & 71.16                                 & -                                    & -                                    \\ \hline
UNITER$_{\text{CC}}$ \cite{UNITER}                       &  71.22 & 22.7                             & 11.8                             \\ \hline
UC$^2$                       &  \textbf{71.48}          & \textbf{34.2} & \textbf{26.8} \\ \hline
\end{tabular}
\caption{\label{tab:vqa} Evaluation results on multilingual VQA task over VQA v2.0 and VG VQA Japanese datasets. We highlight the results for PCATT in {\color[HTML]{0000FF}blue} as they are using different dev/test splits.}


\end{table}
\subsubsection{Evaluation on Multilingual VQA}
For multilingual VQA, our pre-trained model is finetuned and evaluated on the target language for each dataset. Unlike image-text retrieval where the same output layer is shared across different languages, multilingual VQA has different classes of answers for each language, which makes joint training across different languages impossible. We compare our model with state-of-the-art methods without pre-training as well as V+L pre-training methods that use the same pre-training corpus. When evaluating the translate-test baseline $\text{UNITER}_{\text{CC}}$ on VG VQA Japanese dataset, we first finetune it on VQA v2.0~\cite{vqa_v2} with english answer candidates translated from VQA VG Japanese to ensure the same reference is used during evaluation as in UC$^2$. We then use machine translation model to translate the test dataset of VG VQA Japanese to English, and evaluate the finetuned translate-test model using classification accuracy and BLEU. Results are summarized in Table~\ref{tab:vqa}.

On VQA v2.0, our model achieves significant improvement over SOTA task-specific method,
and also outperforms existing monolingual models pre-trained on Conceptual Conception \cite{vilbert, su2020vlbert, UNITER} by an obvious margin. 
On VG VQA Japanese, we finetune our model with a different data split from the original baseline method PCATT proposed in VG VQA Japanese, where we have much less training data than their split (ours: 61K images vs. PCATT: 91K images). Even under this disadvantage in an unfair comparison, our pre-trained model still achieves more than $10\%$ improvement on both accuracy and BLEU over baselines.
Although achieving better performance compared to the task-specific method, the translate test baseline ($\text{UNITER}_{\text{CC}}$) performs much worse than UC$^2$ on the translated VQA VG Japanese dataset. Despite strong performance on the VQA English dataset, the noisiness from the machine translated language would lead to unavoidable degradation especially for tasks like VQA that requires fine-grained level understanding and interpretation on multi-modal context. Hence, building unified cross-linugual cross-modal pre-training model like UC$^2$ is a better solution to directly work on tasks in target languages than a translate-test method.  


\begin{table*}[!]
\centering
\small
\begin{tabular}{l|rrrrrrrr|rrr}
\hline
& \multicolumn{4}{c}{Flickr30K} & \multicolumn{3}{c}{MSCOCO} &  & VQA v2.0 & \multicolumn{2}{c}{VG VQA JA}\\
Pretraiing Objective  & EN  & DE   & FR   & CS    & EN & ZH   & JA    & Meta-Ave  & Test-Dev Acc & Acc & BLEU\\ 
\hline
UC${^2}$ (full model)   & \textbf{88.2} & \textbf{84.5} &  \textbf{83.9} & \textbf{81.2}    &  \textbf{88.1} & \textbf{89.8} & \textbf{87.5} & \textbf{86.2}  & \textbf{71.48} & \textbf{34.2} & \textbf{26.8}\\ 
\hline
\quad\quad -VTLM & 87.5  & 83.6  & 82.4 & 79.6  & 87.7   & 89.2  & 87.2 & 85.3  & 71.45 & 34.1 & 26.7    
\\
\quad\quad -MRTM & 87.6  & 83.7  & 82.0 & 80.0  & 87.9   & 89.4  & 87.4 & 85.4  & 70.93 & 33.5 & 26.4 
\\
\quad\quad-VTLM-MRTM    & 86.8    & 82.9   & 81.3   & 79.3   & 87.5   & 88.9  & 86.7 & 84.8 & 69.94 & 33.4 & 26.4 \\ \hline
\end{tabular}
\caption{\label{tab:itm_abs} Ablation study on pre-training objectives.}

\end{table*}
\begin{table*}[!]
\centering
\small
\begin{tabular}{l|rrrrrrrc}
\hline
& \multicolumn{4}{c}{Flickr30K} & \multicolumn{3}{c}{MSCOCO} & \\
Topology  & EN  & DE   & FR   & CS    & EN & ZH   & JA    & Meta-Ave  \\ 
\hline
UC${^2}$ (Image pivoting)   & \textbf{87.5} & \textbf{83.6} &  \textbf{82.4} & \textbf{79.6}    &  87.7 & \textbf{89.2} & 87.2 & \textbf{85.3}  \\
\hline
UC${^2}$ (English pivoting)    & 86.2    & 81.9   & 80.7   & 77.4   & \textbf{88.1}   & 88.5  & \textbf{87.3} & 84.2 \\ \hline
\end{tabular}
\caption{\label{tab:img_piv_abs} Comparison between the pre-training topology of pivoting on image against pivoting on English.}
\end{table*}

\subsubsection{Ablation Study}~\label{sec:ablate}
\vspace{-3mm}
\paragraph{Effect of Training Objectives} To validate the effectiveness of the proposed pre-training objective MRTM and VTLM, we conduct ablation study to verify their contributions to the model performance. We gradually remove the two proposed training objectives and evaluate these ablated models on our two downstream tasks. 
When finetuning the pre-trained model on the image-text retrieval task, we follow the best experimental setting to train the model on all language data. On VQA task, the model is directly finetuned on the target language data. 

From Table~\ref{tab:itm_abs}, we observe that MRTM has led to significant performance boost on multilingual VQA tasks over the two languages while gaining some incremental improvement on image-text retrieval tasks. VQA requires more fine-grained understanding about connections between language and visual context, therefore benefits more from the cross-modal local alignment captured by MRTM. When introduce VTLM to the pre-training of UC2, we observe similar improvement on image-text retrieval task, but the improvement on VQA VG Japanese is relatively incremental. We suspect the limited help is mainly due to the language gap between English and Japanese captions. Hence it is hard to capture the good alignment between English and Japanese via VTLM. 
\paragraph{Effect of Pivoting on Image}  To validate the effectiveness of image pivoting, we conduct a controlled experiment where the model variant only pivots on English. 
In this setting, we train UC2 with all the pre-training objectives on English Conceptual Caption data, except for VTLM which involves image as one of the pivoting points. To capture the alignment between English and other languages, we train \ucabbr on pairs of captions in two different languages with one language fixed as English. The training objective is translated language modeling adopted from XLM~\cite{lample2019cross}. 
From Table~\ref{tab:img_piv_abs}, we can see that \ucabbr pre-trained by pivoting on image achieves overall better performance in multilingual image-text retrieval task. The advantage is particularly sound when the target language has limited training data. This indicates that the cross-lingual cross-modal representation learned by pivoting on images imbues stronger cross-modal mapping transfer across different languages.




\begin{figure}[h!]
\centering
\includegraphics[width=6.5cm]{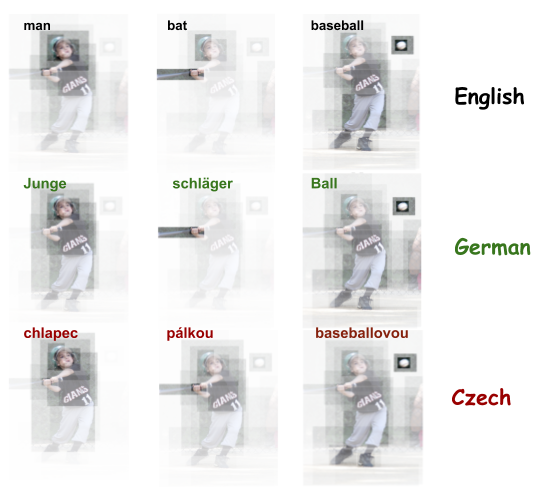}
\vspace{-3mm}
\caption{Visualization of Text-to-Image Attention on aligned words across English, German and Czech (Flickr30K).}
\label{fig:fig2}
\end{figure}

\paragraph{Visualization} To visualize the  cross-lingual cross-modal alignment learned by UC$^2$, we provide examples of text-to-image attention from salient words in multilingual captions to the shared image context. As shown in figure \ref{fig:fig2}, words from different languages that share the same semantic meaning can attend to similar corresponding regions in the image. This shows that while our model can effectively capture cross-modal alignments between regions and words, it also connects different languages by grounding them to similar image regions. 

\section{Conclusion}
We present the first MT-augmented pre-training model UC$^2$ that pivots primarily on images and complementary on English to learn cross-lingual cross-modal representation from large scale of multilingual image-to-text pairs. We propose two new pre-training tasks that facilitate our model to capture better alignment between vision and different languages. Our model achieves the new state-of-art performance on two  mainstream multilingual V+L tasks and demonstrate strong cross-lingual transfer capability. 
For future work, we will continue exploring this topic and expanding the framework to include more families of languages. As more benchmarks \cite{Wang_2019_ICCV, HOW2,bagher-zadeh-etal-2020-cmu} on multilingual video-text pairs become available, we are interested in enhancing the grounding between vision and language by leveraging the temporal information from videos.

{\small
\bibliographystyle{ieee_fullname}
\bibliography{cvpr}
}

\clearpage
\appendix
\section{Appendix} \label{sec:appendix}
In this supplementary materials, we present the implementation details for downstream task finetuning (Section \ref{sec: downstream}) and more ablation study on the proposed pre-training objectives (Section \ref{sec: more ablation}). 
\subsection{Downstream Tasks Details} \label{sec: downstream}
\paragraph{Multilingual Image-Text Retrieval} 
During fine-tuning, we train and evaluate the pre-trained UC$^2$ on Multi30K \cite{multi30k, multi30k_extension_1, multi30k_extension_2} and MSCOCO \cite{MSCOCO, MSCOCO_JA, MSCOCO_ZH}. When we fine-tune UC$^2$ on both datasets, we use batch size of $40$ and sample $2$ negative image-text pairs for each sampled positive image-text pair. The pre-trained model is optimized by the Adam Optimizer with the learning rate set to $1e-4$ and  a linear warm-up for the first $10\%$ of fine-tuning. For Cross-Lingual zero-shot setting, the pre-trained UC$^2$ is fine-tuned on English-only training data for 30K steps. For All-Language setting, we train UC$^2$ on all the training data in all languages for 50K steps. The finetuning is run on 8 Nvidia V100 GPUs. 

\paragraph{Multilingual VQA}
When we fine-tune the pre-trained model on VQA, the output layer is formatted to output a probability distribution over a set of predefined answers. To train the pre-trained model on VQA, we apply a binary cross-entropy loss and optimize the objective with Adam optimizer. On VQA v2.0 \cite{vqa_v2}, we set batch size to $10240$ and the learning rate as $2e-5$, and the model is trained for 7K steps. On VQA VG Japanese \cite{vqa_ja}, the model is trained for 5K steps with the batch of $5120$ and the learning rate of $8e-5$. The weight decay for the fine-tuning on both datasets is set to $0.01$. The fine-tuning for VQA is run on 4 Nvidia V100 GPUs.

\begin{table}[h!]
\centering
\small
\begin{tabular}{l|r|rrr}
\hline
Pre-training tasks  & ITR Meta-Ave  & VQA EN & VQA JA \\ 
\hline
ITM+MLM+MRC & 85.1 & 70.60 & 33.4 \\ \hline
ITM+MLM+MRTM & \textbf{85.3}& \textbf{71.45}& \textbf{34.1}\\ \hline

\end{tabular}
\caption{\label{tab:more_abs} Direct ablation on comparison between the proposed MRTM and the MRC. The presentation of the result is simplified to only include the Meta-Average for the mutilingual image-text retrieval over both Multi30K and MSCOCO, the accuracy on VQA v2.0 test-dev split (referred as VQA EN), and the accuracy on VQA VG Japanese (referred as VQA JA). }
\end{table}

\begin{figure}[h!]
\centering
\includegraphics[width=0.45\textwidth]{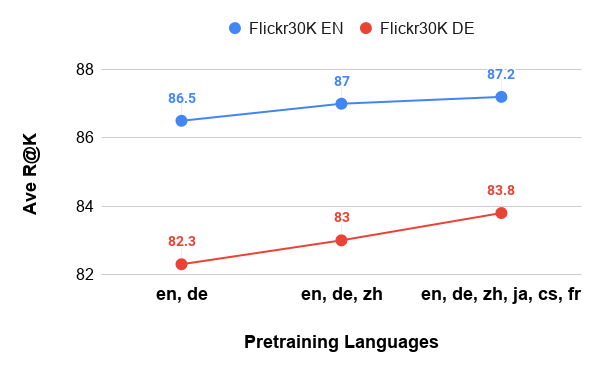}
\vspace{-3mm}
\caption{Comparison of image-text retrieval performance when pre-trained with different groups of languages (average R@K on Flickr30K English and German).} 
\label{fig:1}
\end{figure}

\subsection{Ablation Studies} \label{sec: more ablation}
In this section, we provide more ablation studies on our proposed pre-training objectives that could not fit in the main paper due to space limit. First, we provide evidence to show our proposed MRTM is better than the traditional masked region modeling task MRC for multilingual multi-modal pre-training. Second, we study the impact of number of languages included inpre-training data. Last, we explore the effect of using different pivoting languages in our proposed VTLM task. 

\paragraph{MRTM vs MRC}
For this ablation, we pre-train UC$^2$ with ITM, MLM and MRC and compare the results to the pre-trained UC$^2$ optimized with ITM, MLM and MRTM. The results is summarized in Table. \ref{tab:more_abs}. As shown in table~\ref{tab:more_abs}, compare the pretrained UC$^2$ that employs the traditional task MRC and the one that employs our proposed MRTM, we can see that the performance on the image-text retrieval task are similar, but MRTM leads to marginal improvement on the multilingual VQA tasks. This observation is consistent with our hypothesis that the proposed MRTM augments the local alignment between image regions and the words in different languages which benefits downstream tasks that rely on region-level recognition and reasoning. 

\paragraph{Effect of Pre-training languages}
As we use machine translation models to expand the pre-training corpus, theoretically, we can have as many languages as needed. We conduct further experiments to verify the impact of number of languages included in pre-training data. We create three variants of pre-training corpus, where the number of languages are 2, 3, and 6, respectively.\footnote{For fair comparison, we constraint the training time to be the same with different pre-training corpus.} Every corpus contains English and German. We add Chinese to construct the corpus with 3 languages, and the corpus with 6 languages contains all the languages used to pre-train our full model. The pre-trained models are evaluated on image-text retrieval task in English and German, by finetuning on target language.

Figure~\ref{fig:1} shows that when the number of pre-training languages increases, the performance on image-text retrieval on different languages (English and German)
slightly improves. This result demonstrates that cross-lingual cross-modal pre-training can effectively leverage different languages to learn stronger vision-to-monolingual-sentence alignment. Meanwhile, as we maintain the same pre-training epochs for all three experiments, we also observe that the benefit of multilingual V+L pre-training is compensating for the reduced training time allocated to each language. Although more comprehensive analysis in future study can help us better understand the trade-off between language capacity and performance on downstream tasks, our observation to some extent still suggests that our model is scalable to pre-training on a large corpus with many languages within a reasonable time frame.

\paragraph{Effect of Pivoting Language in VTLM} We also conducted a controlled experiment to learn the effect of different pivoting languages in VTLM for the multi-lingual multi-modal pre-training. In this controlled experiment, we pre-train UC2 with all the objectives but change the pivoting language in VTLM from English to Chinese. When we evaluate the pre-trained model on the multilingual image-text retrieval task, the meta-ave score for the pre-trained model with VTLM pivoted on Chinese is dropped from 86.2 to 85.5. This to some extent suggest that English is a more optimal pivoting language to learn the cross-lingual cross-modal shared representation space. Another potential reason for the limited performance is due to the noiseness in the pre-trained Chinese captions gained via automatic machine translation. To gain more solid conclusion to determine the optimal pivoting language, more comprehensive experiments need to be conducted in the future work. 

\end{document}